\documentclass[conference]{IEEEtran}
\IEEEoverridecommandlockouts
\usepackage{cite}
\usepackage{amsmath,amssymb,amsfonts}
\newtheorem{criterion}{Criterion}
\usepackage{algorithmic}
\usepackage{graphicx}
\usepackage{physics}
\usepackage{textcomp}
\usepackage{url}
\usepackage{xcolor}
\def\BibTeX{{\rm B\kern-.05em{\sc i\kern-.025em b}\kern-.08em
    T\kern-.1667em\lower.7ex\hbox{E}\kern-.125emX}}
    \usepackage{newtxtext,newtxmath}
\begin{document}

\title{How can we trust opaque systems? Criteria for robust explanations in XAI\\
\thanks{© 2025 IEEE.  Personal use of this material is permitted.  Permission from IEEE must be obtained for all other uses, in any current or future media, including reprinting/republishing this material for advertising or promotional purposes, creating new collective works, for resale or redistribution to servers or lists, or reuse of any copyrighted component of this work in other works. 

The research for this paper was generously funded by the German Research Foundation (DFG Grant 508844757), as part of Emmy Noether group \textit{UDNN: Scientific Understanding and Deep Neural Networks}.}
}

\author{\IEEEauthorblockN{Annika Schuster}
\IEEEauthorblockA{\textit{Institute for Philosophy and Political Science} \\
\textit{TU Dortmund}\\
Dortmund, Germany \\
annika.schuster[at]tu-dortmund.de}
\and
\IEEEauthorblockN{Florian J. Boge}
\IEEEauthorblockA{\textit{Institute for Philosophy and Political Science} \\
\textit{TU Dortmund}\\
Dortmund, Germany \\
florian-johannes.boge[at]tu-dortmund.de}
}

\maketitle

\begin{abstract}
Deep learning (DL) algorithms are becoming ubiquitous in everyday life and in scientific research. However, the price we pay for their impressively accurate predictions is significant: their inner workings are notoriously opaque – it is unknown to laypeople and researchers alike what features of the data a DL system focuses on and how it ultimately succeeds in predicting correct outputs. A necessary criterion for trustworthy explanations is that they should reflect the relevant processes the algorithms’ predictions are based on. The field of eXplainable Artificial Intelligence (XAI) presents promising methods to create such explanations. But recent reviews about their performance offer reasons for skepticism. As we will argue, a good criterion for trustworthiness is explanatory robustness: different XAI methods produce the same explanations in comparable contexts. However, in some instances, all methods may give the same, but still wrong, explanation. We therefore argue that in addition to explanatory robustness (ER), a prior requirement of explanation method robustness (EMR) has to be fulfilled by every XAI method. Conversely, the robustness of an individual method is in itself insufficient for trustworthiness. In what follows, we develop and formalize criteria for ER as well as EMR, providing a framework for explaining and establishing trust in DL algorithms. We also highlight interesting application cases and outline directions for future work.
\end{abstract}


\section{Introduction}
Machine Learning systems are nowadays employed throughout science and society. In high-stakes contexts, there is a fear that their employment, in particular using deep learning (DL), bears underestimated, even catastrophic risks \cite{Bengio2023AI}. Part of the reason is that DL is notoriously opaque: It is unknown to laypeople and researchers alike what features of the data a DL system focuses on or how it ultimately succeeds in predicting correct outputs \cite{Boge2022Two}, \cite{Creel2020Transparency},  \cite{Sullivan2022Understanding}. Extensive testing and validation have been argued to partly screen off the relevance of understanding these systems’ detailed workings  \cite{Duran2021Who}, but it is unclear how to judge when the evidence is sufficient, and sufficiently diverse, so that a given model can be trusted. Phenomena such as adversarials  \cite{Goodfellow2018Defense},  \cite{Szegedy2014Intriguing} or clever hans predictions  \cite{Lapuschkin2019Unmasking} suggest that we should remain interested in understanding what the system does and finds. These concerns are addressed by the flourishing field of eXplainable AI (XAI).
Unfortunately, there is some reason for pessimism about XAI  \cite{Ghassemi2021false}: Very often, it is questionable whether individual explanations of DL systems are faithful representations of what the system does, because the methods rely on post hoc interpretations rather than directly reflecting the internal workings of the DL model. For instance, techniques like saliency maps or feature importance rankings can highlight correlations between input features and predictions but may fail to capture the complex, nonlinear relationships that the model actually uses for decision-making. This discrepancy in part arises because such methods typically approximate or simplify the model's behavior, making them susceptible to errors or biases. Moreover, many explanation methods are model-agnostic, meaning they aim to provide insights without direct access to the model's architecture or parameters. While this broad applicability is useful, it also means that the explanations are derived from external observations of the model's outputs rather than from its internal computations. As a result, these explanations may reflect correlations in the data or artifacts of the explanation method itself, rather than the true reasoning of the model. XAI explanations thus may provide an illusion of understanding without truly capturing the mechanisms driving the model's behavior.

How to deal with this situation, in light of the high societal stakes in such contexts as decision making or medical practice, and the high epistemic stakes in contexts of scientific discovery? A natural property to look for as a means for increasing trustworthiness is robustness: Following Levins'  \cite{Levins1966Strategy} famous dictum that the ``truth is the intersection of independent lies'', robustness analysis safeguards against an experimental result or a model-prediction being due to external factors that do not lie in the phenomenon of interest itself, by leveraging several different methods that are not prone to the same kinds of error. Similarly, by probing whether a method yields consistent results under relevant variations of its application, it is possible to ensure that single successes are not just a fluke  \cite{Karaca2022Two},  \cite{Staley2004Robust}. In this way, researchers can reasonably increase the trust they legitimately invest in their methods and results  \cite{Boge2024Why}.

Traditionally, philosophical accounts of RA have focused on the idealizing assumptions contained in different models and the insensitivity of theoretical results to these assumptions  \cite{Levins1966Strategy},  \cite{Weisberg2006Robustness}. Others have scrutinized the insensitivity of experimental results to the peculiarities of specific experiments \cite{Karaca2022Two}, \cite{Staley2004Robust}, \cite{Boge2024Why}. Only recently, the robustness of DL models, has caught philosophers’ attention \cite{Freiesleben2023Beyond}, \cite{Grote2024Reliability}.

In the technical literature, robustness is typically identified with the insensitivity of DL models' outputs to (specific kinds of) changes in the input. A special class of inputs that has taken center stage in the technical literature are so-called `adversarials': inputs that contain what seems like a small, even human-imperceptible alteration, but have a drastic impact on an DL model's outputs. In \cite{BogeDeep}, a more encompassing notion of robustness in DL is suggested, since DL anomaly-detectors need to be sensitive to subtle change. According to them, DL robustness should be interpreted as the insensitivity of a DL model to certain kinds of changes in input, architecture and overall design, or even the training regimen and testing conditions, with details depending on the specific context. \cite{Freiesleben2023Beyond} offer a similarly general definition, wherein robustness in DL means that a certain target will not undergo changes greater than allowed by some specific tolerance, if a certain modifier is changed.\footnote{They interpret robustness as a causal concept, since models need to be implemented on a physical computer (also \cite{Boge2019Why}, \cite{Boge2020How}). However, robustness analysis can be applied to models \textit{in abstracto} \cite{Weisberg2006Robustness}, where the relevant connection may be a deductive relation between assumption and result. It would seem strange to say that changing assumptions `causes' the results to change.}  

We are here not concerned with the robustness of DL results or models themselves, though, but with explanations of their workings. Hence, a reasonable request is explanatory robustness (ER): That the use of several XAI methods yields consistent results. However, as we shall argue in what follows, ER is not enough: several XAI methods may yield consistent results, yet there could be reason to think that they are all wrong nevertheless. Thus, there is a prior need for Explanatory-Method Robustness (EMR): Several distinct but relevantly related explanations by means of a single XAI method should be consistent.\footnote{\cite{Karaca2022Two} proposes a similar distinction between result robustness and procedural robustness.} In effect, since explanations are needed for trustworthy DL and EMR is needed for trustworthy explanations, EMR comes out as a precondition for trustworthy DL on our analysis.

The article is structured as follows: We begin with some considerations about the explanations XAI provides, their limits and potential (section 2). Then, we introduce the notions of explanatory robustness (section 3) and explanatory method robustness (section 4). In section 5, we provide evidence that our criteria are relevant in practice, by identifying studies that (tacitly) make use of them. In section 6, we discuss our findings and in section 7, we consider limitations of the present study and highlight directions for future work.

\section{Explanations in Philosophy of Science and XAI}
This section summarizes the main positions on scientific explanations from philosophy of science and then discusses the status-quo of XAI explanations with regard to the previously introduced explanation types, the abstractiveness-faithfulness-trade-off and the ground-truth problem.

\subsection{Scientific explanations}

Explanations are answers to why-questions. They consist of a phenomenon to be explained, typically termed the explanandum, and an explanans that provides the explanation according to some specific relation between the two. In philosophy of science, several accounts of scientific explanation have been proposed. The classical, deductive-nomological (D-N) model \cite{Hempel1948Studies} argues that the explanandum has to be deduced from general laws together with the conditions of the concrete situation to be explained (explanans). For example, you can deduce the length of a flagpole’s shadow from the flagpole’s length and the angle of the sun, together with the laws of trigonometry (tangent function). Later accounts have criticized the symmetry of this relation (one could, for example, explain the flagpole’s length from its shadow’s length and the angle of the sun in the same vein), the lack of laws to deduce explanations from and the fact that explanatorily irrelevant factors can be included in D-N-explanations without making a difference (one could, for example, explain the flagpole’s length from its shadow’s length together with irrelevant statements about weather conditions). 

In part as a reaction to these deficits of the D-N account (see \cite{Salmon1989Four}), many other \emph{types} of explanation have been identified, such as causal \cite{spirtes2000, pearl2000}, causal-mechanistic \cite{machamer}, functionalist \cite{cummins}, or statistical-relevance explanations \cite{salmon}. Furthermore, pragmatic and unificationist approaches to explanation propose general conditions that a set of propositions needs to satisfy to count as an explanation (for an overview, see \cite{sep-scientific-explanation}). 

In more detail, Salmon's account of statistical relevance explanation \cite{salmon} cites a specific partition of a statistical population to provide those factors that are maximally informative about the probabilistic structure of some effect. It is thus thoroughly statistical, and not necessarily causal.

 In contrast, the causal-mechanistic account focuses on establishing causal mechanisms that bring about the explanandum as explanantia. Newer variants of mechanistic accounts stand in explicit opposition to functionalist or dynamicist accounts which state that ``mathematical and computational models can explain a phenomenon without embracing commitments about the causal mechanisms that produce, underlie, or maintain it'' \cite[p.~605]{Kaplan2011Explanatory}, and justify the explanatory power of their explanations in providing fine-grained descriptions as well as accurate predictions. Mechanistic explanations, instead, have the following goal:
\begin{quote}
To explain the phenomenon, the model must in addition reveal the causal structure of the mechanism. This will involve describing the underlying component parts, their relevant properties and activities, and how they are organized together causally, spatially, temporally, and hierarchically.\cite[p.~605]{Kaplan2011Explanatory}
\end{quote}
They highlight the importance of a mapping between the elements in the model and elements in the real-world mechanism for explanatory power. 

Kitcher’s unificationist account of explanation proposes that explanantia in general follow argumentative patterns that can be instantiated by many different phenomena \cite{Kitcher1981Explanatory}. Thus subsuming different phenomena under common patterns is what promotes understanding, on Kitcher's account. Pragmatist accounts instead highlight the relevance of contextual factors in an explanation: 
\begin{quote} 
The discussion of explanation went wrong at the very beginning when explanation was conceived of as a relation like description: a relation between a theory and a fact. Really, it is a three-term relation between theory, fact, and context. [...] An explanation is an answer [...] what is requested differs from context to context.\cite[p.~156]{Van1980Scientific}
\end{quote} 

For pragmatists, thus, there is no ideal explanation over all contexts. Instead, contextual factors decide which propositions can fulfill the relevant explanatory goals.

This short, pointed review of explanation types in philosophy of science shows that depending on the requirements on the explanation, different techniques can be justified, as they all have their (dis)advantages and it can be argued that different explanation types are best for different explanatory goals. To evaluate whether an explanation answers the posed why-question satisfactorily, it makes sense to explore the possibility space of answers. If, for example, the question is ``Why is there a rainbow in the sky?'', the D-N-explanation would deduce it from the laws of optics and atmospheric conditions, while a causal-mechanistic explanation would focus on the relations between the events that constitute the rainbow (sun rays pass through the atmosphere, entering rain droplets present in the air that refract them, etc.). The epistemic value of each explanation might partly lie in the subsumption under some kind of common pattern, as promoted by the unificationist account, and the  context may determine which pattern is of greatest interest, according to the pragmatist account. We submit that these factors -- the relevance of different types of explanations for different purposes and the understanding-promotion power of recognizing similar patterns -- are relevant also in the context of XAI.

\subsection{XAI explanations}

This subsection discusses which kind of explananda and explanantia are found in the context of DL models, how the aforementioned explanation types fit into the current XAI-practice, and, finally, the implications of the ground-truth problem for XAI explanations. It ends with a short summary.

\paragraph{Explanandum, explanans and the abstractiveness-faithfulness-trade-off}

First, it is useful to think about what plays the role of explanans and explanandum in XAI explanations. Depending on whether an XAI technique is local or global, the explanandum or phenomenon to be explained are either single predictions or the whole DL model. Notice that the focus of most XAI techniques is not to explain phenomena in the world with the help of DL models, but to illuminate the models' inner processes: ``‘explanation’ [...] refers to an understanding of how a model works, as opposed to an explanation of how the world works.'' \cite[p.~206]{Rudin2019Stop}. However, we might not always even be interested in the inner workings of the DL model at all, but rather looking for structures in the input data that the DL model uses but that are not immediately visible to humans: ``An interpretation may prove informative even without shedding light on a model’s inner workings. [...] The real goal might be to explore the underlying structure of the data [...].'' \cite[p.~11]{Lipton2018mythos} Both the DL model as well as structures in the input data (local or global), or even features inferred from these \cite{Boge2022Two}, are thus valid explananda for XAI explanations.

The explanans in XAI explanations is difficult to grasp. The main sources of opacity in DL models are their size, which makes their parameters and calculations ungraspable to humans all at once \cite{Sgaard2023On}, and their instrumentality  \cite{Boge2022Two}. S\o{}gaard illustrates the opacity due to size as follows: 
\begin{quote}
Humans cannot be expected to familiarize themselves with $n$ parameters for sufficiently high $n$, and they obviously cannot hold this many parameters in short-term memory in an attempt to reconstruct inference dynamics.\cite[p.~228]{Sgaard2023On}
\end{quote}
 Boge \cite{Boge2022Two} emphasizes that it is opaque what is being learned by DL models, and that this is a separate problem from understanding the model. In addition, he points to the lack of \textit{ab initio} content in their parameters, which makes them purely instrumental as models:
\begin{quote}
The elements of the model that could be used as representations are weights, biases, and activations. But [...] these are merely adjustable parameters devoid of content, not representations that help us conceptualize some underlying mechanisms by facilitating visualization, qualitative mathematical reasoning, or causal inference.\cite[p.~54-55]{Boge2022Two}
\end{quote}
Explanantia in XAI will thus make DL models or their acquired contents (what they have learned) more graspable -- by aggregating and simplifying the information they contain and by assigning meaningful interpretations to its instrumental parameters, or by emphasizing overall properties of their learning and functioning that render them understandable by displaying a distinctive, recognizable pattern. 

There is a natural trade-off between the complexity and informativeness of an explanation, also called the abstractiveness-faithfulness-trade-off \cite{Sgaard2023On}. The most faithful explanation is ultimately a complete description of the DL model, but it is too complex for us to understand. The more graspable an explanation becomes, the more it abstracts away from details. The goal is thus to find a solution to what S\o{}gaard terms a minimization problem:
\begin{quote}
Our objective thus becomes a weighted minimization problem, minimizing a weighted sum of the discrepancy between our approximation and the original model, and the size of our approximation.\cite[p.~234]{Sgaard2023On} 
\end{quote}
Which level of abstraction is appropriate is context-dependent. When researchers debug a model, or medical AI is analyzed to draw conclusions about signs of specific illnesses, more faithful explanatory information is of interest. But when a model user is trying to decide whether or not to trust a DL model’s prediction, a more abstract and thus more graspable explanation would generally be preferred.

\paragraph{Types of explanations} While it is disputable that there even is an encompassing one-one mapping between philosophical types of explanation, and while XAI methods generally need to be supported by relevant background knowledge to even provide explanations properly-so-called \cite{bogemosig}, it is interesting nevertheless to investigate the relation between philosophical accounts of explanation and XAI-`explanations'.

Erasmus et al.\ \cite{erasmus2021} argue that DL models are D-N-explainable because one can cite the configuration of the network as a lawlike-connection and the input as a special condition to derive the classification, which is then ipso facto D-N explained. However, we are skeptical that it is correct to think of the network as instantiating a law \cite{prasetya2022}. Furthermore, this account does not make contact with practices in XAI. More promising are recent approaches to symbolic regression involving a DL model's hidden layers \cite{wetzel2025}, which could explain the law-like structures discovered by the model. 

An explanation technique for DL models that can, under certain circumstances, provide mechanistic insight is SHAP, which provides information on how features contribute to either a single model decision or globally. The perturbations used to arrive at feature attributions do not generally establish causal relations for the user. They can, however, sometimes be used to identify potential causal factors that can then be empirically confirmed \cite{bajorath}.

Some advanced explanation techniques seem to take the requirements of causal-mechanistic explanations to heart. For example, mechanistic interpretability research \cite{Olah2020Zoom}, \cite{Kastner2024Explaining} focuses on illuminating functions of different parts of the architecture in DL model and counterfactual analysis \cite{Wachter2017Counterfactual}, \cite{Mothilal2020Explaining} provides users with answers to what-if-things-had-been-different questions, which can provide causal information in case they can be associated with interventions \cite{woodward1997explanation}.

The unificationist and pragmatist accounts of explanation do not propose distinct types of explanation, but argue, respectively, that explanation promotes understanding by unifying a broad range of phenomena under a small stack of argument patterns \cite{Kitcher1981Explanatory}, and that any explanation will depend on pragmatic factors for its success, such as the contrast class of potential alternative answers to a why-question \cite{Van1980Scientific}. As argued in \cite{paez2019pragmatic}, such pragmatic elements are endemic to all XAI models. Furthermore, XAI explanations of general learning mechanisms such as the information bottleneck framework \cite{shwartz2017opening} clearly strive for unification, although the class of DL models subject to that framework's constraints has been shown to be limited \cite{saxe2019}.
 
As for the type of explanation, it has been shown \cite{raz} that the information bottleneck framework corresponds to a generalization of statistical relevance explanations à la Salmon \cite{salmon}. Other explanation techniques that do not faithfully represent the causal structure of a DL model fit functionalist explanations \cite{cummins}. For instance, certain approaches to symbolic regression that leverage DL for data compression \cite{udrescu} are not faithful to the causal structure of the DL model, though they might be faithful to the causal structure of the underlying domain. When it comes to explaining the DL model, they thus highlight the model's functional capacity to reproduce a physical connection, instead of providing evidence that the model exploits the underlying equations or that its outputs are causally influenced by relevant features. 

In view of the variety of properties of XAI explanations, we propose to take a strongly pluralist stance: Not only are there different types of explanation for different phenomena; many phenomena can be explained with different explanation techniques, where depending on context one might be preferable over the other. This allows to contextualize XAI methods with respect to the concrete explanatory problem at hand. 

This type of explanatory pluralism is defended in \cite{Bokulich2018Searching}, where it is argued that different, rival explanations of some phenomenon may be valuable if they either highlight the phenomenon's causal structure, or alternatively allow to unify it with apparently similar but causally distinct phenomena: 
\begin{quote}
[T]here can be more than one scientifically acceptable explanation for a given phenomenon at a time. [...] This pluralism, rather than revealing some sort of shortcoming in our understanding [...], is in fact one of its great strengths. [p. 17]
\end{quote}

There is an abundance of XAI techniques that cannot all be discussed within the scope of this article. We will thus focus the remaining discussion exemplarily on saliency maps, LIME and SHAP. The proposed criteria are, however, intended to be general and should be applicable to all XAI techniques.

\paragraph{The ground-truth problem}

A fundamental difficulty in assessing XAI explanation quality is that we typically lack the ground truth – there is no definitive standard against which we can measure whether an XAI method's explanation is correct. As there are to date no generally agreed upon definitions of interpretability and explainability, and in addition these concepts seem to have very different characteristics in different situations (depending e.g., on the goal of the explanation and the model architecture), the benchmarks employed to evaluate XAI techniques often cannot serve as objective standards:
\begin{quote}
Regardless of the explanation purpose, and with little conceptual motivation, they formally define properties that they are optimizing their explanations for. Other explanation techniques (often designed for completely different applications and optimized for distinct desiderata) are then compared according to their own standards. In this form, benchmarks lose their justification; they become advertisement space rather than an objective standard for comparison. \cite[p.~6]{Freiesleben2023Dear}
\end{quote}
In addition, new techniques are developed at a very fast pace, leaving little space for reflection on possible standards. This point is also noted in \cite[p.~3]{Alangari2023Exploring}: ``There is an urgent need in the ML interpretability research field to focus more on comparing and assessing the existing explanation methods instead of continuing to create new methods.'' These quotes show that an in-depth analysis of the current XAI methods is needed.
\paragraph{Discussion}

The main problem in XAI explanations, in addition to contextuality and the related abstractiveness-faithfulness-trade-off, is thus a general way to evaluate them. In what follows, we show that explanatory method robustness in addition to explanatory robustness introduces a plausible way of doing so. While we argue that our criteria should be seen as a necessary condition for trustworthy AI, domain-specific benchmarks that take the various explanatory needs in different contexts into account would, in our view, be a valuable addition to clarify and evaluate the explanation goals in context.

\section{Explanatory Robustness in XAI}

As defined above, explanatory robustness (ER) in the context of XAI refers to the fact that several different means for explaining something about an AI system yield consistent results. Thus we propose the following criterion:

\begin{criterion}[Explanatory Robustness (ER)]
An explanation in XAI can only be considered robust if different XAI methods with the same goal (e.g., explaining the system’s overall functioning or an individual decision) produce
\begin{enumerate}
\item the same explanations for the same DL model, or input-output pair, respectively;
\item distinct explanations for distinct models, or input-output pairs, respectively.
\end{enumerate}
\end{criterion}

More precisely, let $z_i=\ev{x_i,y_i}$ input output pairs of an DL model, $f$, and $X, \hat{X}$ the outputs of two XAI methods, $F, \hat{F}$, where both can either be global, $X=X(f)$, or local, $X = X(z_i)$. Then ER can be formally defined as follows:

\begin{align}
& d(X(z_i), \phi(\hat{X}(z_i))) <\varepsilon \tag{\text{local ER-1}} \\
& d(X(f), \phi(\hat{X}(f)))<\varepsilon', \tag{\text{global ER-1}}
\end{align}
where $\varepsilon, \varepsilon'$ are contextually determined tolerances, $\phi$ is a suitable transformation that allows comparison between the outputs of $F$ and $\hat{F}$, and $d$ is a suitable metric. The existence and definition of $\phi$ is non-trivial, but we suggest that it can usually be found. For instance, Shap-CAM \cite{zheng2022shap} allows for comparison of Shapley values to heatmaps on images, by generating heatmaps from Shapley values.

Furthermore, it has long been known that there can be instances where several methods agree though they should not, and that this spoils the credibility of a given robustness analysis: It suggests that they agree for the wrong reasons. This phenomenon has been called `discriminant validity' in the context of biological modeling \cite{Campbell1959Convergent}, and has been characterized as exposing spurious robustness on examples from high-energy physics \cite{Karaca2022Two}, \cite{Staley2004Robust}. We thus propose an additional set of conditions for ER, with the same effect:

\begin{align}
& d'(z_i, z_j) >\gamma \Rightarrow d(X(z_i), \phi(\hat{X}(z_j))) >\epsilon, \tag{\text{local ER-2}} \\
& D(f, f')> \gamma' \Rightarrow d(X(f), \phi(\hat{X}(f'))) >\epsilon', \tag{\text{global ER-2}}
\end{align}
where $\gamma, \gamma', \epsilon, \epsilon'$ are again contextually determined, and $d'$ is a metric on the space $\mathcal{X}\times \mathcal{Y}$ of input-output pairs. $d'$ may actually be relaxed to a divergence for comparing distributions, or it could involve some prior transformation of inputs and outputs respectively, so as to allow for a more meaningful comparison (say, in terms of semantic properties of input-output pairs, not bare, na\"ive proximity in the input and output-space, respectively). We can also allow it to have two sets of arguments and to be vector-valued, $d':\mathcal{X}\times\mathcal{Y}\rightarrow \mathbb{R}^2\cup\qty{\infty}^2$, so that differences in $\mathcal{X}$ and $\mathcal{Y}$ may be evaluated independently and possibly by different metrics or divergences. 

The definition of $D$ is similarly non-trivial, but since the softmax or other final sigmoid layer allows the interpretation of DL models as distributions, we suggtes that a divergence such as the Kullback-Leibler or a symmetricized metric like the Shannon-Jensen may usually do the work (although more sophisticated measures for comparison that also pay attention to architectural details might sometimes be required). Finally, since at least sometimes, different input-output pairs may credibly receive the same explanation, ER-2 may be relaxed as follows:

\begin{align}
    &\mathrm{Pr}[d(X(z_i), \phi(\hat{X}(z_j)))<\Delta| d'(z_i, z_j)>\gamma]<\lambda  \tag{\text{local ER-2'}},\\
    &\mathrm{Pr}[d(X(f), \phi(\hat{X}(f')))<\Delta'| D(f, f')>\gamma]<\lambda'  \tag{\text{global ER-2'}},
\end{align}
where $\mathrm{Pr}[\cdot | \cdot ]$ is a conditional probability function, and $\Delta, \Delta', \lambda, \lambda'$ are also contextually determined. We note here that conditions ER-1 could be modified in obviously analogous ways so as to also demand (a high probability of) similar explanations for merely \emph{similar} models, or similar input-output pairs, but we forego further discussion here. 
 
The possibility of ER has an important precondition; namely, that the different methods have the same explanatory goal. It would not make sense to consult both a global SHAP explanation and a local LIME explanation in the same robustness analysis. In order for a robustness analysis with the goal of establishing ER to even potentially succeed, all methods considered must either target the functionality of the DL model (global ER), or the structures in the data relevant for prediction (local ER) – or whatever other feature or property we might be interested in. They cannot each target a different aspect. 

Why is ER important? ER embodies the basic requirement of all robustness analyses, that a result should be attainable via several different but relevantly related routes so as to not be explainable as an artifact of an individual method \cite{schupbach}. Such effects are, however, reported for XAI methods like SHAP and LIME: In some studies \cite{Neely2021Order}, their explanations have been shown to rarely agree with each other. 

Prima facie, methods like SHAP, LIME, or saliency maps are subject to different sources of error and have different strengths. Conversely, if one does get consistent results from several distinctly created saliency maps, SHAP and LIME, an explanation based on the overlap might be said to satisfy ER: Several distinct methods, prone to different kinds of error, suggest the same explanation, making it seem unlikely that the explanation is artificial. If this is supplemented by the fact that these methods do not agree `blindly' (ER-2), it increases credibility further. However, since the ground-truth is generally lacking and these methods do often disagree with each other, we currently have insufficient grounds to trust their outputs. More importantly, we suggest that it is not even sufficient to just use several random XAI techniques to validate an explanation, as the agreement might still be a fluke. Instead, each of the methods has to be robust in itself, a property we call \textit{explanatory method robustness}. 

\section{Explanation Method Robustness}
Due to DL models’ blackbox nature, explanatory robustness is not enough to speak of a robust explanation. A fundamental difficulty in assessing explanation quality is that we lack the ground truth – there is no definitive standard against which we can measure whether an XAI method's explanation is correct. This uncertainty complicates efforts to determine whether an explanation faithfully represents the model’s decision-making process.
We therefore suggest that, to mitigate this problem, explanatory method robustness (EMR) has to be fulfilled in addition to ER, where EMR is defined as follows:  

\begin{criterion}[Explanation Method Robustness (EMR)]
An XAI method can only be considered robust if it outputs:
\begin{enumerate}
    \item similar explanations for similar models, or input-output pairs, respectively;
    \item distinct explanations for distinct models, or input-output pairs, respectively.
\end{enumerate}
\end{criterion}
Formally:
\begin{align}
  d'(z_i, z_j)<\varepsilon & \Rightarrow  d(X(z_i), X(z_j))<\delta \tag{\text{EMR-1}} \\
    d'(z_i, z_j)>\varepsilon' &\Rightarrow  d(X(z_i), X(z_j))>\delta'  \tag{\text{EMR-2}},
\end{align}
where again, the tolerances are conventionally determined. However, since at least sometimes, different input-output pairs may credibly receive the same explanation, EMR-2 may be relaxed as follows:

\begin{equation}
    \mathrm{Pr}[d(X(z_i), X(z_j))<\Delta| d'(z_i, z_j)>\varepsilon']<\lambda  \tag{\text{EMR-2'}},
\end{equation}
with analogous meanings as above. 

EMR has thus been formulated only locally, w.r.t.\ input-output pairs, but it also admits of a global formulation, if the method can explain the model globally. Thus, a saliency map should explain two very similarly trained convolutional nets similarly, and should presumably explain a transformer and a convolutional net differently. Formally:

\begin{align}
  D(f, f')<\varepsilon & \Rightarrow  d(X(f), X(f'))<\delta \tag{\text{global EMR-1}} \\
    D(f, f')>\varepsilon' &\Rightarrow  d(X(f), X(f'))>\delta'  \tag{\text{global EMR-2}},
\end{align}

or, with the probabilistic weakening as above,
\begin{equation}
    \mathrm{Pr}[d(X(f), X(f'))<\Delta'| D(f, f')>\varepsilon']<\lambda'  \tag{\text{global EMR-2'}}.
\end{equation}

\section{Application cases and connections to other work}
Condition EMR-1 is sometimes identified with XAI robustness tout court \cite{DBLP:journals/corr/abs-1806-08049}, but we believe it falls short of exhausting a credible notion: robustness analysis must compare the agreement of certain results, either from different methods or across relevantly related results from one single method. As has been shown in dedicated case studies of the robustness of methods in experimental and observational sciences \cite{Campbell1959Convergent, staley2004}, there can be spurious agreement in such an analysis, and this is an indication that their agreement may be spurious overall. Furthermore, if several methods satisfy EMR-1 and EMR-2 but yield mutually incompatible results in the sense of defying ER, we cannot trust the explanations they each deliver. 

The process we propose is depicted schematically in Figure 1. After EMR has been established for all methods that are relevant for the given explanation, individual explanations for the prediction or model in question are compared in a robustness analysis, at the end of which the robustness of the explanation can be evaluated.

\begin{figure}[!t]
  \centering
  \includegraphics[width=\columnwidth]{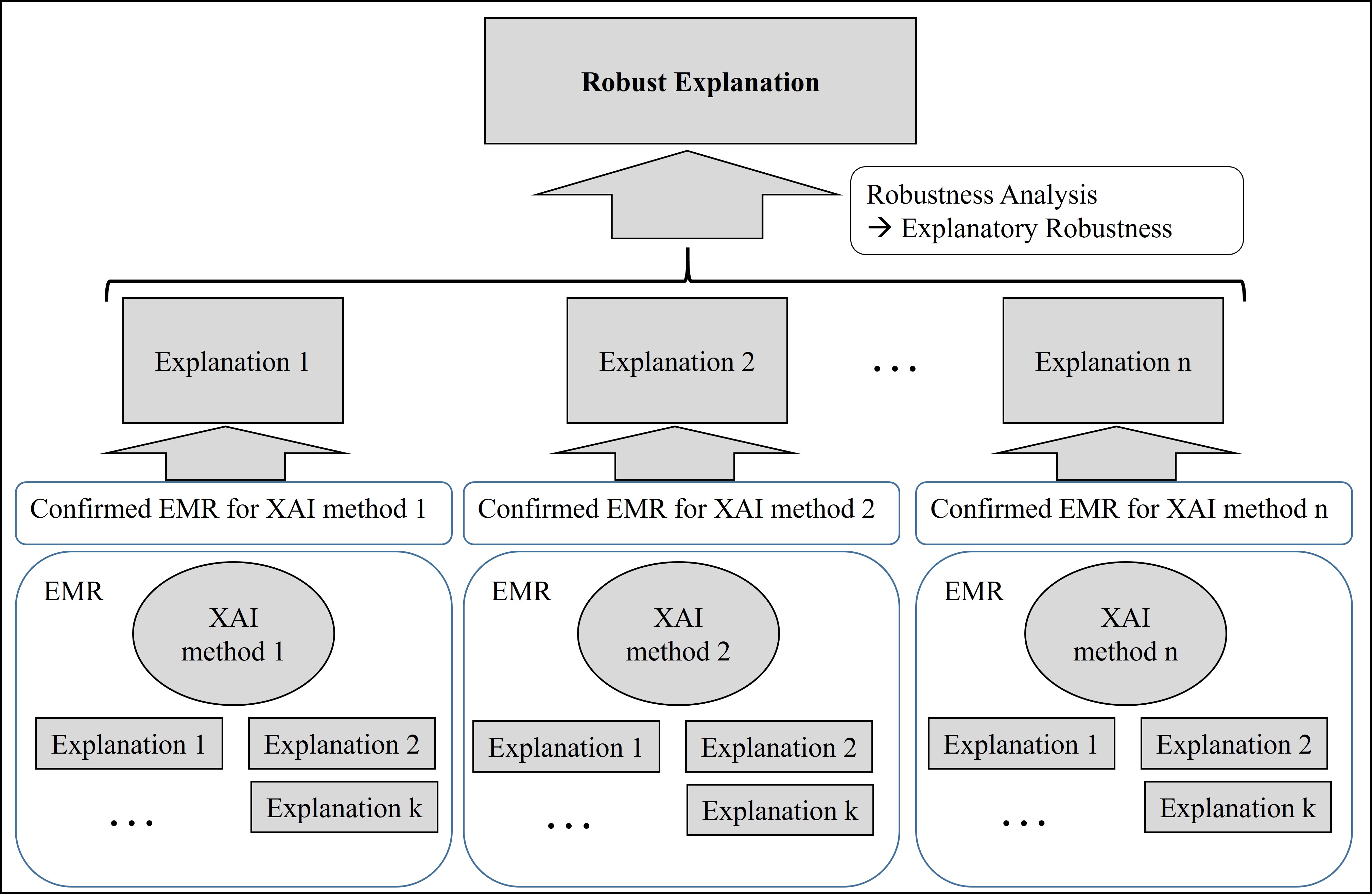}
  \caption{Schematic depiction of the proposed robustness analysis making use of Explanatory Method Robustness and Explanatory Robustness.}
  \label{fig:example1}
\end{figure}

EMR-1 reflects the explanation’s robustness to noise. It is similar to  the definition of ML robustness in \cite{Freiesleben2023Beyond}, wherein the term means that a certain target will not show changes greater than a predefined tolerance, if a certain modifier is changed. [41] investigate robustness in the sense of EMR-1 for many different XAI methods. They found that the criterion was not fully satisfied by any of those methods, but that gradient-based methods were more robust than model-agnostic methods. EMR-2 reflects the explanation’s sensitivity to variations in model-behavior. While in rare cases, distinct input-output pairs might receive the same explanation (EMR-2'), it is generally implausible to assign similar explanations to different results. If exceptions to this rule are known, EMR-2'  (and analogously ER-2') might be further relaxed by letting $\lambda^{(')}=\lambda^{(')}(z_i, z_j)$ and specifying paired patches or regions of the space $\mathcal{X}\times\mathcal{Y}$ for which explanations should likely agree despite the remoteness of the $z_i, z_j$ or $f, f'$.

A major challenge in XAI is the presence of adversarial examples – cases where imperceptible input changes lead to drastic prediction shifts. If an XAI method satisfies EMR-1 but not EMR-2, it might assign the same explanation to cases that are similar in the input space $\mathcal{X}$ but clearly distinct in the output space $\mathcal{Y}$, concealing the model’s vulnerability. Conversely, failing EMR-1 while satisfying EMR-2 could lead to erratic explanations that overstate sensitivity to minor variations. Thus, EMR provides a diagnostic framework for ensuring that explanations accurately reflect both stability and meaningful sensitivity in model behavior, which could complement adversarial robustness efforts.

Despite not being explicitly embraced as gold standards for XAI explanations, criteria like ER and EMR have already been used to criticize XAI methods. As we explained above, EMR-1 is the most widespread condition. EMR-2, on the other hand, has been employed in extant criticisms of saliency maps: \cite{Adebayo2018Sanity} compared the outputs of several distinctly created saliency maps on trained models to their outputs randomly initialized, untrained networks with same architecture. Since the results were strikingly similar in each case, this signals a violation of global EMR-2 for most saliency maps. 

\cite{Kindermans2017Un} present results that show that a constant shift in the input data causes several kinds of saliency maps to attribute incorrectly, even though the network prediction is identical. This constitutes a violation of EMR-1, assuming $d$ compares inputs and outputs independently and the $\mathcal{X}$-component is chosen as a divergence that compares pixel-distributions modulo colour-information. Pairs with shifted inputs would then be identified by $d$, but still explained differently by the saliency map. On the other hand, Rudin \cite{Rudin2019Stop} presents a case wherein two radically different predictions are both explained by highly similar saliency maps, thus yielding a violation of (local) EMR-2. As was pointed out above, ER is presupposed in comparisons of methods like SHAP and LIME \cite{Neely2021Order}, where it has been shown to fail on certain instances. 

The lack of fulfilment of EMR and ER for popular XAI methods exhibited in these studies shows the importance of widely employing these notions. As no solution to the ground-truth problem exists, the consistency of explanations guaranteed by EMR and ER is in general the only available means for increasing the trustworthiness of explanations. While consistency alone does not ensure correctness \cite{Raz2023}, explanations that fail EMR or ER cannot be considered reliable, as they vary unpredictably across similar cases or fail to distinguish meaningfully different ones.

\section{Discussion}

The explanations presented in XAI suffer from a basic flaw: there are no commonly agreed upon standards against which they can be compared. Instead, a large number of different metrics and benchmarks are usually employed to argue for their usefulness. Trustworthy explanations are important, in particular in high-stakes contexts, and require general standards for credibility. Our criteria for robust XAI, which, as we have shown, are not fulfilled by several popular XAI methods, make an important step in this direction. Only if methods can at least be shown to be consistent in themselves and amongst each other, can they be reasonably trusted. 

Domain-specific benchmarks that take the various explanatory needs in different contexts into account would, in our view, be a valuable addition to clarify the explanation goals in context. A step back from human evaluations of explanation and towards more rigorous standards for testing the methods in different conditions is thus an important step towards trustworthy DL despite opacity.

\section{Limitations}
This is a purely conceptual study, hence no novel empirical results, providing evidence for the utility of the notions proposed, have been presented. However, we have offered formal definitions that allow operationalisation and will aid empirical studies in future work. Furthermore, we have provided evidence that existing studies exhibit failures of one or other of our criteria, and that this is recognized to impair trustworthiness. We thus submit that the present paper will aid future systematic approaches to establishing DL trustworthiness. Notably, the cited studies only show that \emph{failures} of either ER or EMR impair trustworthiness. What remains to  be determined by future empirical work is how strongly the \emph{fulfilment} of ER and EMR correlates with credible results, as could be determined by external validation metrics in benchmark studies where a correct explanation is known.

\bibliographystyle{IEEEtran}
\bibliography{ref-extracts.bib}
\end{document}